# Atlas-based automated detection of swim bladder in Medaka embryo

Diane Genest[1,2], Marc Léonard[2], Jean Cousty[1], Noémie De Crozé[2] and Hugues Talbot[1,3]

[1] Université Paris-Est, Laboratoire d'Informatique Gaspard Monge, CNRS, ENPC, ESIEE Paris, UPEM, F-93162 Noisy-le-Grand
[2] L'OREAL Research & Innovation, F-93600 Aulnay-sous-Bois
[3] CentraleSupélec, Université Paris-Saclay, équipe OPIS-INRIA, F-91190, Gif-sur-Yvette

**Abstract.** Fish embryo models are increasingly being used both for the assessment of chemicals efficacy and potential toxicity. This article proposes a methodology to automatically detect the swim bladder on 2D images of Medaka fish embryos seen either in dorsal view or in lateral view. After embryo segmentation and for each studied orientation, the method builds an atlas of a healthy embryo. This atlas is then used to define the region of interest and to guide the swim bladder segmentation with a discrete globally optimal active contour. Descriptors are subsequently designed from this segmentation. An automated random forest classifier is built from these descriptors in order to classify embryos with and without a swim bladder. The proposed method is assessed on a dataset of 261 images, containing 202 embryos with a swim bladder (where 196 are in dorsal view and 6 are in lateral view) and 59 without (where 43 are in dorsal view and 16 are in lateral view). We obtain an average precision rate of 95% in the total dataset following 5-fold cross-validation.

## Introduction

Fish embryo models are widely used in both environmental and human toxicology and for the assessment of chemicals efficacy [1]. According to the international animal welfare regulations, fish embryos are ethically acceptable models for the development of toxicological assays, with the complexity of a complete organism [2,3]. As an aquatic organism, fish embryos can provide relevant information about the environmental impact of chemicals. The study of early developmental stage malformations provides relevant information on developmental toxicity of chemicals [4,5,6]. The absence of swim bladder, an internal gas-filled organ that allows the embryo to control its buoyancy is one of the most sensitive marker of developmental toxicity [7,8]. In this article, we focus on the automated detection of the swim bladder on 2D images of 9 days post fertilization (dpf) Medaka embryos (*Oryzias Latipes*).

The proper detection of the swim bladder depends on the orientation of fish embryos. This implies to manually place the fish embryo before the image acquisition, which is tedious and time consuming. To overcome this difficulty, we developed a method capable of detecting the swim bladder regardless of the fish embryo position. Here, a methodology based on morphological operators is proposed to automatically detect the swim bladder on 2D images of fish embryos seen either in dorsal view or in lateral view [9,10]. After a pre-processing step that includes the automated determination of the embryo orientation [11], the methodology consists of the generation of an atlas representative of a healthy embryo [12,13,14], of swim bladder segmentation using this atlas, of descriptors calculation and of embryos classification according to the presence or absence of a swim bladder. Using this method on a dataset of 261 images, we obtain an accuracy of 95%. The successive steps of the method are represented in Fig *2*.

Section 1 introduces the swim bladder localization step, including the atlas generation in 1.1, the identification, using this atlas, of a region of interest (ROI) for the search of the swim bladder, and the swim bladder segmentation by a globally optimal active contour algorithm applied on this ROI in 1.2 [15]. Then Section 2 describes how the segmented shape is characterized with intensity and morphological descriptors extraction. Finally, Section 3 presents the process of embryos classification by a random forest classifier according to the presence or absence of a swim bladder [16].

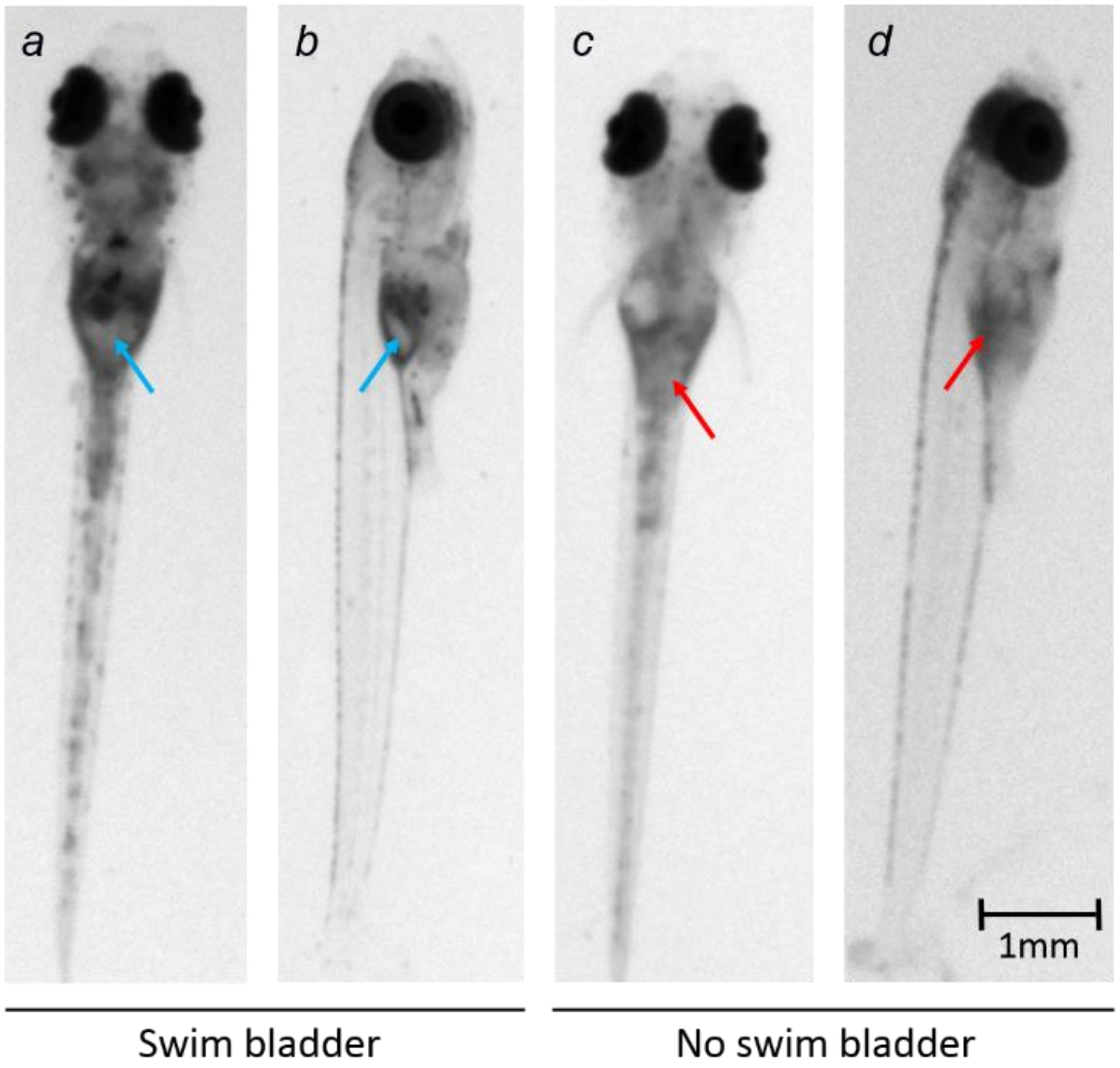

**Fig 1.** Medaka embryos with or without swim bladder. The blue arrow indicates the swim bladder location for embryos with a swim bladder (a and b) and the red arrow indicates the location where a swim bladder should be present for embryos without a swim bladder (c and d). a and c: embryos seen in dorsal view; b and d: embryos seen in lateral view.

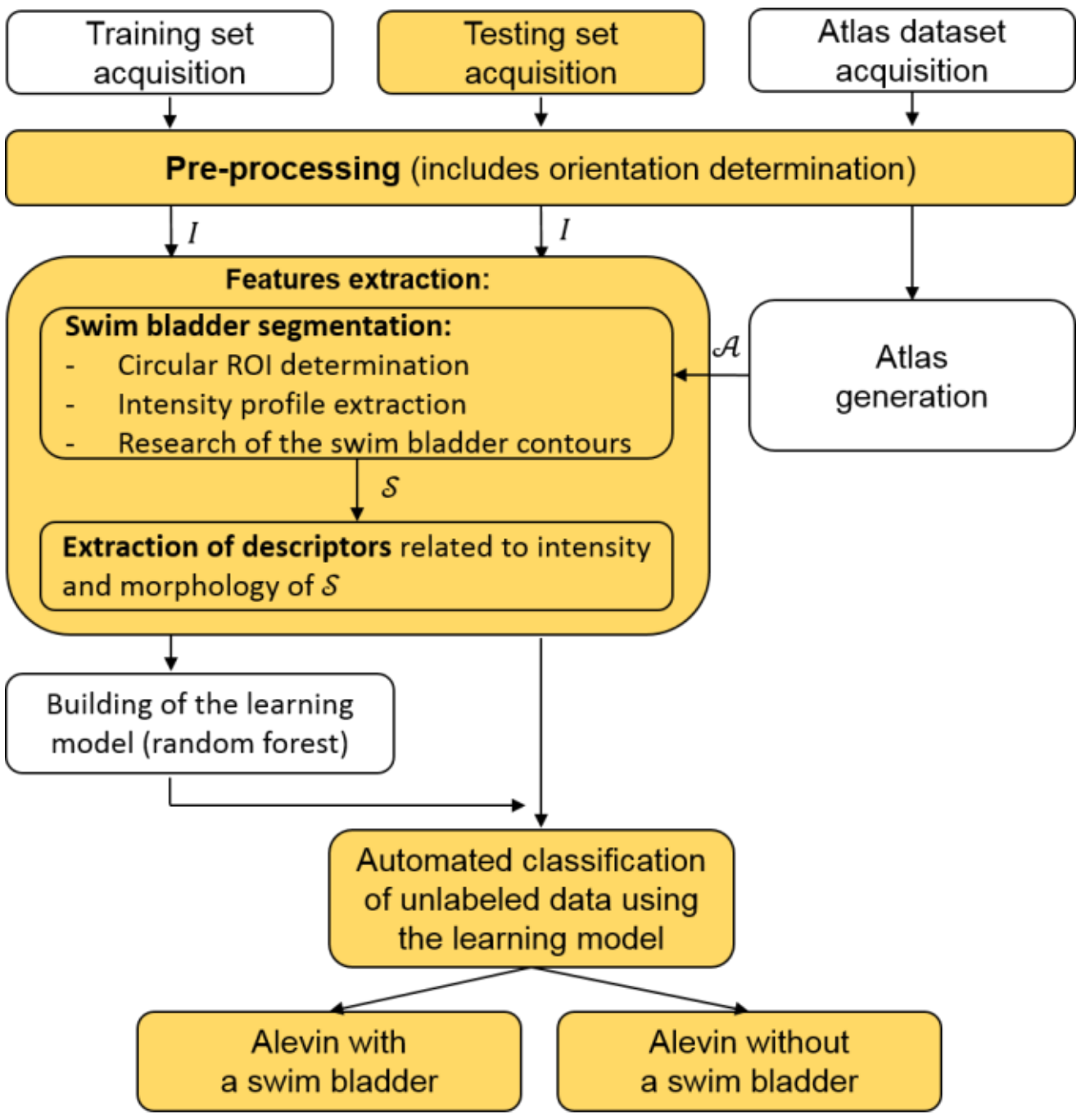

**Fig 2.** Flowchart of the swim bladder detection assay

# 1 Swim bladder localization

As a swim bladder can be present or absent in embryos images, the aim of this part is to find the most probable contours of a swim bladder in such images. For practical reasons, the swim bladder segmentation method is described in this part with respect to the embryo orientation that appears the most frequently in our database, which is the dorsal view. However, this methodology is adaptable to embryos seen in lateral view with minimal parameter changes. As we also have developed a method to automatically determine the orientation of the embryo based on orientation related descriptors, it is feasible to fully automate the swim bladder detection process.

## 1.1 Swim bladder atlas generation

The objective here is to build a median image $I_{med}$ representative of a typical healthy embryo with respect to the diversity of embryos that exists in experimental conditions, and a probability function $p_{sb}$ defined on the ensemble of $I_{med}$ pixels coordinates and that represents the likelihood of each pixel of the represented image to belong to the swim bladder. In the following, such a pair $(I_{med}, p_{sb})$ is called an atlas of the swim bladder for Medaka embryo images and will be denoted by $\mathcal{A}$.

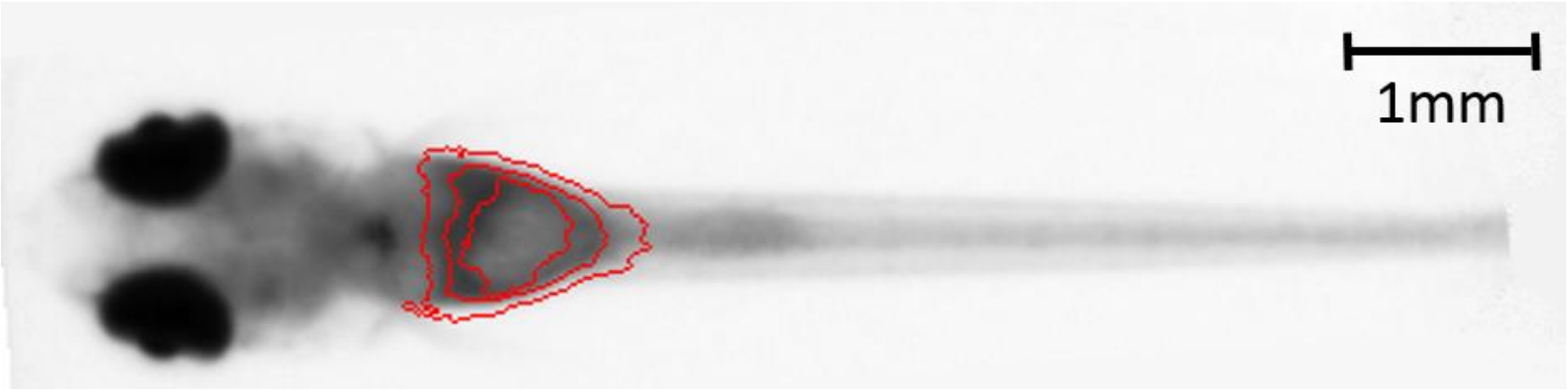

**Fig 3.** Atlas $\mathcal{A} = (I_{med}, p_{sb})$ obtained for fish embryos seen in dorsal view. The three red lines show the iso-contours that delimit the areas where the pixels have a probability equal to 1, to 0,5, and to 0,05 to belong to the swim bladder.

In order to build the atlas, $n$ images of healthy embryos are selected and their swim bladder are manually segmented. Among these images, one is randomly chosen as being the fixed reference image $I_F$. Then, the $n - 1$ remaining images, called the moving images hereafter, are aligned on $I_F$ by applying an affine image registration algorithm. It consists of finding, for each moving image, an affine transformation that minimizes a similarity measure between the fixed image and the transformed moving image [17,18]. The similarity measure used here is the mutual information. This maximizes the measure of the mutual dependence between the pixel intensity distributions of the fixed and the moving images, defined by Thévenaz and Unser as:

$$MI(\mu; I_F, I_M) = \sum_m \sum_f p(f, m; \mu) \, log_2 \left(\frac{p(f,m;\mu)}{p_F(f) p_M(m;\mu)}\right) \quad (1)$$

where $m$ and $f$ are the intensities of the fixed and moving image respectively, $p$ is the discrete joint probability, and $p_F$ and $p_M$ are the marginal discrete probabilities of the fixed image $I_F$ and the moving image $I_M$ [19,20]. The multiresolution affine registration algorithm from the elastix toolbox is used to perform this process [21]. The median $I_{med}$ of the $n$ registered images is calculated (Fig 3). For each moving image $I_M$, the optimal transformation $\mu$ is also applied to the corresponding manual segmentation of the swim bladder. The average $I_{av}$ of the $n$ registered swim bladder segmentations is calculated. We define as the probability function $p_{sb}$ the mapping that, to each pixel of coordinates $(x, y)$ of a new image $I$ with the same dimensions as $I_{med}$, associates the value $I_{av}(x, y)$. The atlas $\mathcal{A} = (I_{med}, p_{sb})$ will then be used in order to identify, in a new image $I$, the ROI where to search the swim bladder.

## 1.2 Swim bladder segmentation

After registering the atlas on a new image $I$, we observe that the registered atlas does not segment the contour of the swim bladder with precision, as shown in Fig 4. As our methodology relies on the characterization of the swim bladder most probable contours to distinguish embryos with and without a swim bladder, we need to obtain a more accurate delineation of the swim bladder, when it is present.

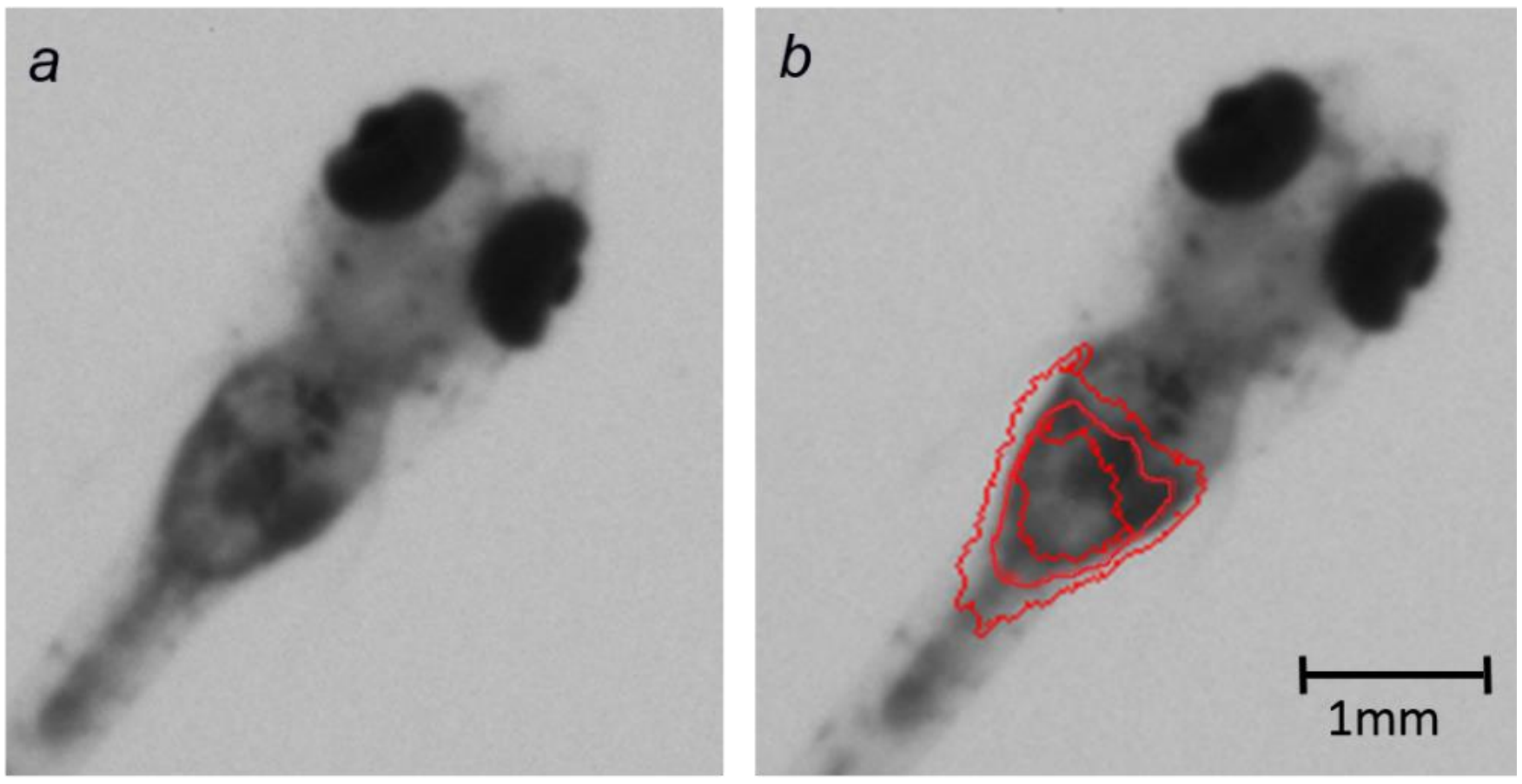

**Fig 4.** Projection of the atlas $\mathcal{A}$ on an embryo image. The three red lines delimit the areas where pixels have a probability equal to 1, to 0.5 and to 0.05 to belong to the swim bladder according to the probability function $p_{sb}$ of the atlas.

**ROI localization.** The atlas $\mathcal{A} = (I_{med}, p_{sb})$ is used on an embryo image $I$ in order to identify the region of interest in which the swim bladder will be searched. To this aim, we search for the transformation $\mu'$ to apply to $I_{med}$ that optimally registers $I_{med}$ to the analyzed image $I$. We apply the same affine registration process as described in Section 1.1. This transformation $\mu'$ is then applied to the definition ensemble of $p_{sb}$, leading to a new transformed probability function $p'_{sb}$. We compute the binary mask $M$ that contains the pixels of $I$ where the probability of finding the swim bladder, if a swim bladder is present, is equal to 1. The barycenter of this binary image $M$ is extracted and considered as the center $C$ of the ROI. The image ROI is then defined as the circle of center $C$ and of diameter 40 pixels, experimentally determined (Fig 5).

**Extraction of the swim bladder most probable contours.** As shown in Fig 1, the swim bladder is characterized by a high contrast between dark contours and light inner part that cannot be too small. By contrast, embryos without a swim bladder present a homogeneous body in the location where the swim bladder should be present. For this reason, the swim bladder detection method relies on the determination of a circular shortest path extracted from the image ROI represented in a dual polar frame defined as follow. Considering ROI of center $C = (\mathrm{x}_C, \mathrm{y}_C)$ and of radius $r$ in the primal frame of the image $I$, we define its associated representation in a dual polar frame $\mathrm{ROI}_d$, as the image that is the concatenation of all ROI radial sections, starting from a radial section $s_1$ that is perpendicular to the embryo skeleton obtained during pre-processing.

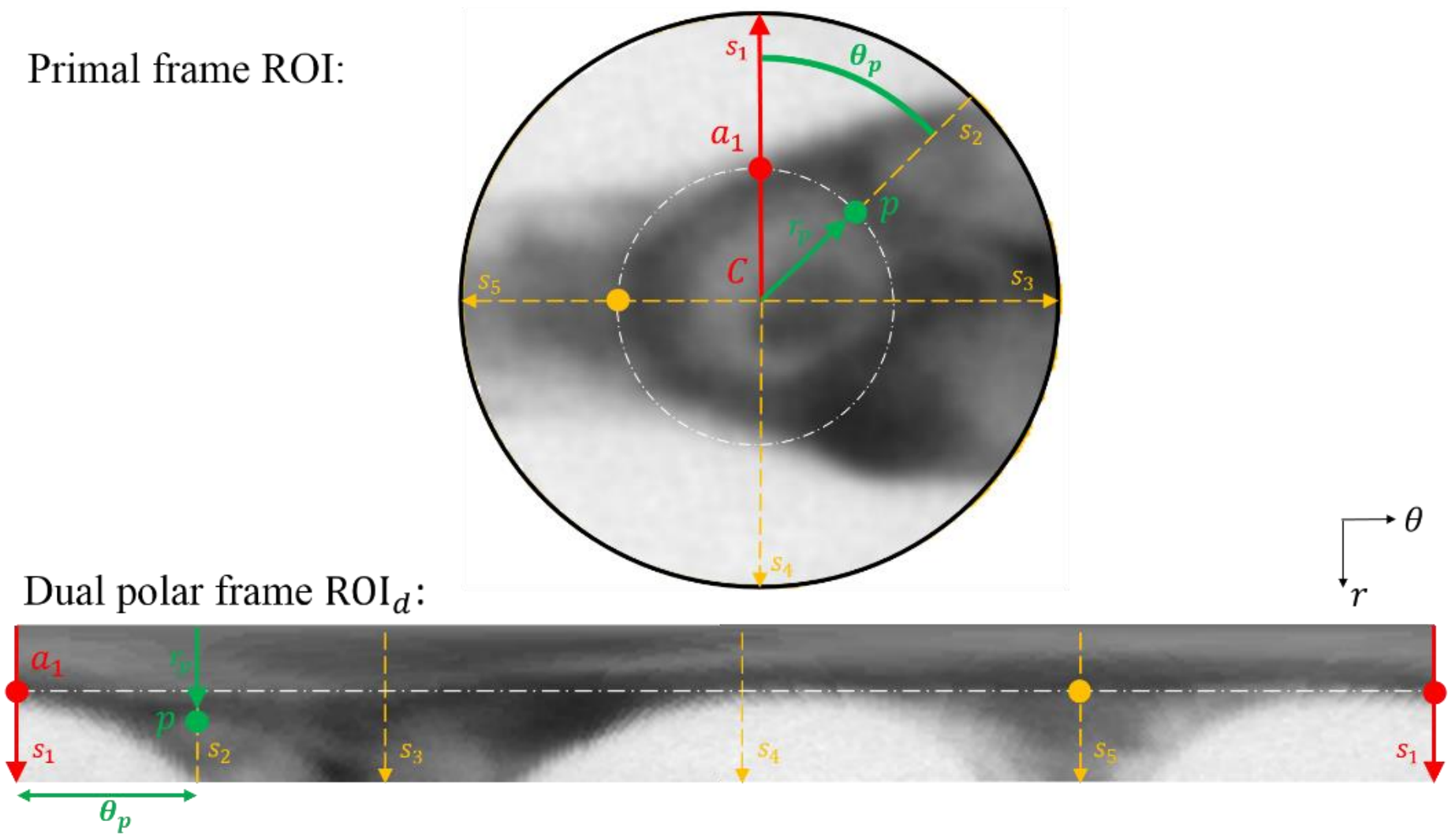

**Fig 5.** Representation of the image ROI in the primal frame and of its associated image $ROI_d$ in the dual polar frame.

Each radial section $s_\theta$ of the image ROI, defined by an angle $\theta$ in degrees from the initial section, is then precisely the $\theta^{th}$ column of the dual image representation $\mathrm{ROI}_d$ (Fig 5). Hence, to each pixel $p = (\theta_p, r_p)$ in $\mathrm{ROI}_d$, we associate the intensity $c(p)$ of the pixel $p' = (r_p \cos\theta_p, r_p \sin\theta_p)$ in the primal image ROI.

We then consider circular shortest paths in the image $\mathrm{ROI}_d$, *i.e.*, paths corresponding to contours of minimum energy in the primal image [15][22]. For this purpose, the image $\mathrm{ROI}_d$ is equipped with a directed graph such that a pair $(a, b)$ of two pixels of $\mathrm{ROI_d}$ is a directed arc if $a_1 = b_1 - 1$ and $|a_2 - b_2| \leq 1$, where $a = (a_1, a_2)$ and $b = (b_1, b_2)$. In this graph, a *circular path* is a sequence $(p_0, \dots, p_l)$ of pixels of $\mathrm{ROI_d}$ such that:

- for any $i$ in $\{1, \dots, l\}$, the pair $(p_{i-1}, p_i)$ is an arc;
- the first coordinate of $p_0$ is equal to 0;
- the first coordinate of $p_l$ is equal to 360 (*i.e.* the maximal possible value); and
- the second coordinate of $p_0$ and of $p_l$ are the same.

The *energy cost* $EC(\pi)$ of a circular path $\pi = (p_0, \dots, p_l)$ is defined as the sum of the intensities of the pixels in the path: $EC(\pi) = \sum_{i \in \{0,\dots,l\}} c(p_i)$. A circular path $\pi = (p_0, \dots, p_l)$ is called *optimal* whenever the energy cost of $\pi$ is less than or equal to the energy cost of any circular path from $p_0$ to $p_l$. Circular optimal path can be computed with any graph shortest path algorithm such as the one of Dijkstra [22, 23].

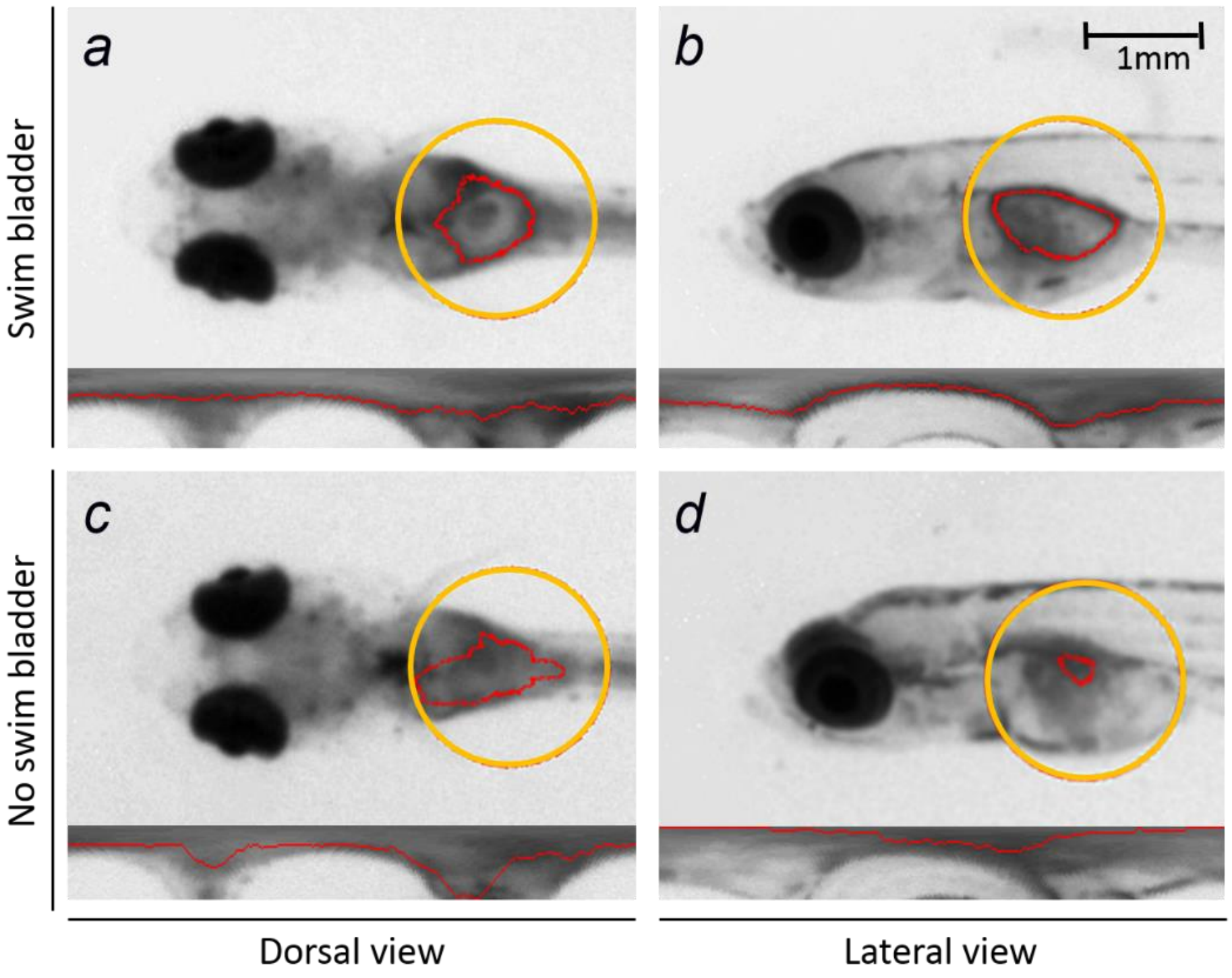

**Fig 6.** Swim bladder segmentation results on the primal frame of the image and associated shortest path in the dual polar frame. The yellow circle delimits the $ROI$ in which the red line shows the contour of the segmented shape $\mathcal{S}$ in case of embryos with a swim bladder (a and b) and embryos without a swim bladder (c and d). a and c: embryos seen in dorsal view. b and d: embryos in lateral view.

In order to obtain the most probable contour of the swim bladder, we start by selecting the most peripheral local minimum of the first radial section $r_1$, called $a_1$. We also define $rmin$ as the minimal radius of the ROI below which the shortest path must not be searched. It is experimentally set to 10 pixels.We then consider a circular shortest path $\pi$ starting at $a_1$.This circular shortest path found in the image $\text{ROI}_d$, corresponds to a closed contour $\mathcal{S}$ in the image ROI which surrounds the center $C$ and which is of minimal energy. Such optimal contour $\mathcal{S}$ is hereafter referred to as the most probable contour of the swim bladder (Fig 6). This methodology expects to segment the swim bladder if present, or a random part of the embryo body otherwise. Swim bladder characterization will now allow to distinguish both cases.

## 2 Swim bladder characterization

We now need to identify if the segmented shape $\mathcal{S}$ corresponds to a swim bladder or not, basing on descriptors of the swim bladder.

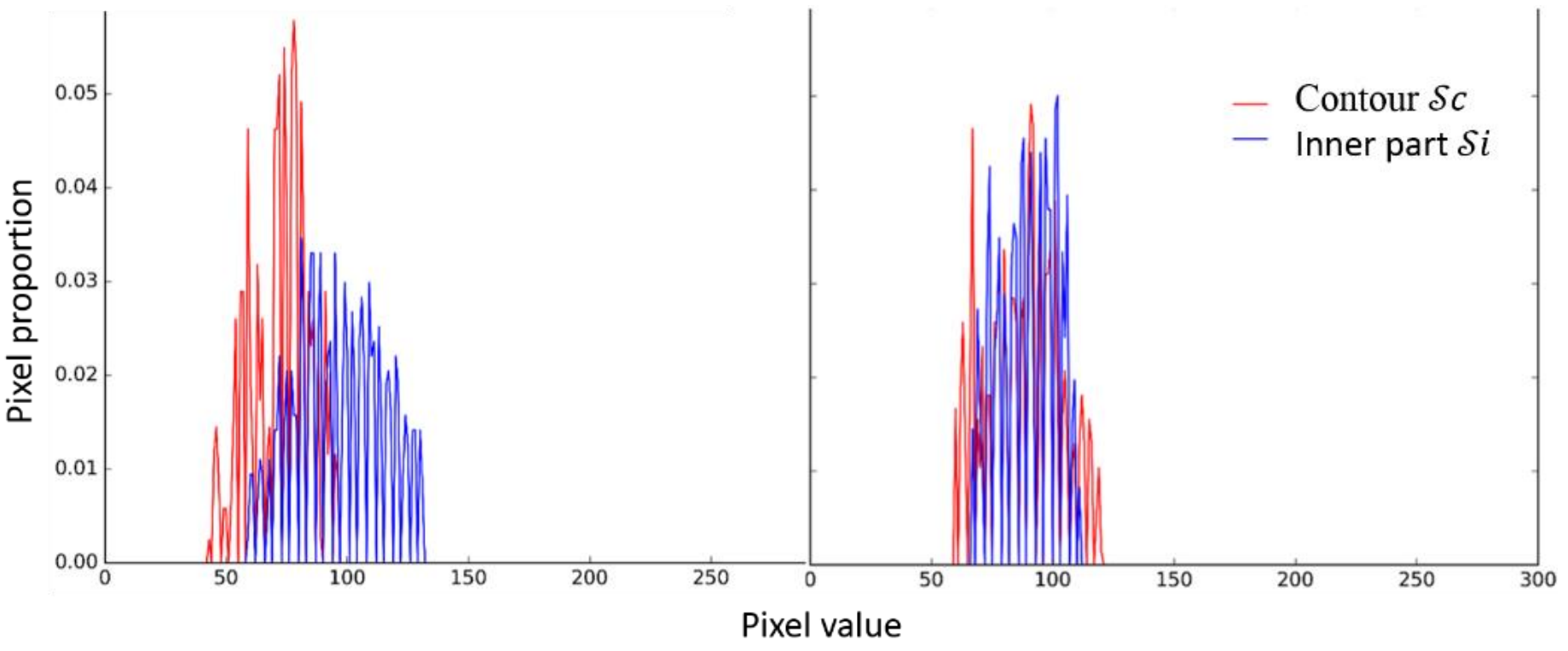

**Fig 7.** Histograms of the inner part $\mathcal{S}i$ and the contour $\mathcal{S}c$ of the segmented shape $\mathcal{S}$.

## 2.1 Intensity descriptors

A swim bladder is characterized by a high contrast between a dark contour and a lighter inner part, contrary to an embryo without any swim bladder that presents a more homogeneous intensity in the delimited shape. On the histograms of both the inner part $\mathcal{S}$i and the contour $\mathcal{S}$c of the segmented shape $\mathcal{S}$, this means that a shift is visible between both distributions of pixel intensities in case of an embryo with a swim bladder (Fig 7). The following intensity-related parameters are extracted from both histograms: the minimal, maximal, average and median intensities, the intensity mode, and their piecewise differences are calculated in order to characterize the contrast between $\mathcal{S}$ inner part and contour. We also measure the two ranges, *i.e.*, the difference between the maximal and the minimal intensities, and the ratio between them. We finally compute the pixel intensity variance of $\mathcal{S}$, and the covering, defined as follow. We calculate the difference between the maximal value of $\mathcal{S}$c and the minimal value of $\mathcal{S}$i in one hand, and the difference between the maximal value of $\mathcal{S}$i and the minimal value of $\mathcal{S}$c on the other hand. The covering is defined as the ratio between both differences. These intensity-related descriptors will then be combined with other descriptors characteristic of the swim bladder morphology which are described in the following section.

## 2.2 Morphological descriptors

In order to characterize the swim bladder shape, the following descriptors are extracted from $\mathcal{S}$. We refer to the *convexity* of a swim bladder, as the set difference between the convex hull of $\mathcal{S}$ and $\mathcal{S}$ itself [24]. We refer to the *concavity* of a swim bladder as the area of the deleted component after an opening of $\mathcal{S}$ by a disk-shape structuring element of size 5. We furthermore consider the elongation of $\mathcal{S}$ defined by $\frac{4 \times area(\mathcal{S})}{perimeter(\mathcal{S})^2}$. All descriptors related to intensity or morphological characterization of $\mathcal{S}$ are summarized in Table 1.

**Table 1.** List of descriptors used for swim bladder characterization.

| Parameter name | Parameter description | Parameter name | Parameter description |
|---|---|---|---|
| Intensity descriptors: | | | |
| $variance_i$ | Variance of pixel intensity | $min_diff$ | $min_{\mathcal{S}i} - min_{\mathcal{S}c}$ |
| $min_i$ | Minimal intensity | $max_diff$ | $max_{\mathcal{S}i} - max_{\mathcal{S}c}$ |
| $max_i$ | Maximal intensity | $av_diff$ | $average_{\mathcal{S}i} - average_{\mathcal{S}c}$ |
| $average_i$ | Average intensity | $mode_diff$ | $mode_{\mathcal{S}i} - mode_{\mathcal{S}c}$ |
| $mode_i$ | Histogram mode | $median_diff$ | $median_{\mathcal{S}i} - median_{\mathcal{S}c}$ |
| $median_i$ | Median intensity | $rrange$ | $\frac{range_{\mathcal{S}i}}{range_{\mathcal{S}o}}$ |
| $range_i$ | $max_i - min_i$ | $covering$ | $\frac{max_{\mathcal{S}c} - min_{\mathcal{S}i}}{max_{\mathcal{S}i} - min_{\mathcal{S}c}}$ |
| | | $min_diff$ | $min_{\mathcal{S}i} - min_{\mathcal{S}c}$ |
| for $i$ is either $\mathcal{S}i$ or $\mathcal{S}c$ | | $max_diff$ | $max_{\mathcal{S}i} - max_{\mathcal{S}c}$ |
| Morphological descriptors: | | | |
| $convexity$ | $convex_hull(\mathcal{S}) - \mathcal{S}$ | $elongation$ | $\frac{4 \times area(\mathcal{S})}{perimeter(\mathcal{S})^2}$ |
| $concavity$ | $area(\gamma_5(\mathcal{S}) - \mathcal{S})$ | | |

# 3 Fish embryos classification

We detail in this section the classification process used to assess the methodology.

## 3.1 Experimental setup

On day 0, fish eggs are placed on a 24-well plate, one egg per well, in 2mL of incubation medium with or without chemicals [25]. The ninth day, embryos are anesthetized by adding Tricaïne to the incubation medium (final concentration: 0,18g/L) and one image of each embryo is recorded. Each embryo is then observed in all possible orientations under a microscope by an expert. The expert annotates each fish embryo as having or not a swim bladder. This observation constitutes the ground truth.

With this protocol, we set up a total database of 287 images of embryos, where 259 are seen in dorsal view (91.5%) and 28 in lateral view (8.5%). A subset of this database is used in order to generate the atlas described in Section 1.1. For reasons linked to the unbalanced proportions of both orientations in our total available database, we select $n = 20$ images of healthy embryos seen in dorsal view for dorsal atlas generation, and $n = 6$ healthy embryos seen in lateral position for lateral atlas generation. The remaining 261 images constitute the validation dataset. Among them, 202 present a swim bladder according to the ground truth, and 59 do not present a swim bladder. In particular, the subset of 202 embryos with a swim bladder is composed of 196 images of fish embryos seen in dorsal view and 6 seen in lateral view. The subset of 59 embryos without a swim bladder is composed of 43 seen in dorsal view and 16 in lateral view.

### 3.2 Classifier description

A random forest classifier is defined with the following parameters [16,26]: the number of estimators is set to 50, the maximal depth to 30, the minimal number of samples required to split an internal node to 5 and the minimal number of samples required to be at a leaf node to 2. The entropy criterion is chosen.

### 3.3 Classification results and discussion

The results of our classification process after a 5-fold cross validation are presented in Table 2 in the form a confusion matrix that shows the distribution between embryos with and without a swim bladder according to the ground truth and to the prediction results. The classifier reaches an accuracy score of 95% in the total dataset. The sensitivity is 90% and the specificity is 96%.

This article aimed to describe a method for the automated detection of the swim bladder in Medaka fish embryos. As a screening test, we assess the accuracy of the method by calculating the overall accuracy, the sensitivity and the specificity. With an overall accuracy of 95% measured, this study reveals the feasibility of an automated detection of the swim bladder from 2D images of Medaka embryos. We also observe a high specificity of 96% and an acceptable sensitivity of 90%, which suggests that the use of this method is accurate for the global screening test that is to be developed. Indeed, the swim bladder detection assay is intended to be a part of a series of morphological abnormalities detection assays whose each specificity needs to be maximized in order to maximize the accuracy of the whole detection assay and not eliminate a too important number of chemicals [11,27]. The sensitivity of each individual test is expected to improve the sensitivity of the whole test.

The experimental protocol of our method presents the advantage to not require the manual positioning of each embryo in the well. After anesthesia, embryos remained in the incubation medium in their well and can have any possible orientation before images are acquired and treated. However, this protocol does not permit to control the orientation . In our database, a disproportion between natural positioning of embryos exists. Without any control on embryo positioning, we obtain highly unbalanced datasets between dorsal and lateral views. In our total dataset of 261 images, we only have 8.5% of embryos seen in lateral view vs. 91.5% of embryos seen in dorsal view. Among the lateral embryos, 90,9% are correctly classified, vs. 95,4% for the dorsal embryos. These results reveal that the presence or the absence of swim bladder can be detected with a satisfying accuracy in embryos images, regardless of their orientation, with an adaptive swim bladder detection method. As we are also able to detect automatically the embryos orientation, that means this process can be fully automated. In a future work, we plan to adapt this method to the intermediary orientations, by performing a 3D reconstruction of a healthy embryo from dorsal and lateral atlases interpolation [28,29].

**Table 2.** Classification results after 5-fold cross validation.

| Prediction: / Ground truth: | Swim bladder | No swim bladder |
|---|---|---|
| Swim bladder | 195 | 7 |
| No swim bladder | 6 | 53 |

## 4 Conclusion

This article proposes a methodology to automatically classify images of Medaka embryos according to the presence or the absence of swim bladder. It relies on the use of a healthy embryo atlas. This atlas is used to define the research area of the swim bladder in the analyzed embryo. The most probable contour of the swim bladder is segmented in this research area and descriptors are extracted to characterize the segmented shape. The final classification using a random forest classifier yields a resulting accuracy of 95% in the total dataset, which is satisfying, in particular with respect to the major advantage of the method of being orientation-independent. In terms of experimental manipulations, being able to automatically analyze embryos regardless to their orientation would save significant time as the protocol would not require neither any manual positioning nor any visual inspection of embryos to detect abnormalities. To increase the precision and analyze intermediary orientations, we plan to generate a 3D atlas.